\documentclass[runningheads]{llncs}
\usepackage{subfiles}
\usepackage{amsmath}
\usepackage{multicol}
\usepackage[skip=2pt,font=small]{caption}
\usepackage[T1]{fontenc}

\usepackage{graphicx}     
\usepackage{subcaption}
\usepackage[misc]{ifsym}

\usepackage{graphicx}
\usepackage{xcolor}
\usepackage{comment}
\begin{document}
\title{Improving the Evaluation and Actionability of Explanation Methods for Multivariate Time Series Classification}

\titlerunning{XAI for Multivariate Time series Classification}
%
\author{Davide Italo Serramazza \Letter \and Thach Le Nguyen \and Georgiana Ifrim}

\tocauthor{Davide Italo Serramazza, Thach Le Nguyen, Georgiana Ifrim}
\toctitle{Improving the Evaluation and Actionability of Explanation Methods for Multivariate Time Series Classification}
\authorrunning{Serramazza et al.} 
%
\institute{School of Computer Science, University College Dublin, Ireland \\
\email{davide.serramazza@ucdconnect.ie}\\
\email{\{thach.lenguyen,georgiana.ifrim\}@ucd.ie}}

\maketitle   
\begin{abstract}
Explanation for Multivariate Time Series Classification (MTSC) is an important topic that is under explored. There are very few quantitative evaluation methodologies and even fewer examples of actionable explanation, where the explanation methods are shown to objectively improve specific computational tasks on time series data. In this paper we focus on analyzing InterpretTime, a recent evaluation methodology for attribution methods applied to MTSC. 
\textcolor{black}{We showcase some significant weaknesses of the original methodology and propose ideas to improve both its accuracy and efficiency.} 
Unlike related work, we go beyond evaluation and also showcase the actionability of the produced explainer ranking, by using the best attribution methods for the task of channel selection in MTSC. We find that perturbation-based methods such as SHAP and Feature Ablation work well across a set of datasets, classifiers and tasks and outperform gradient-based methods. We apply the best ranked explainers to channel selection for MTSC and show significant data size reduction and improved classifier accuracy.

\end{abstract}
\section{Introduction}

 Machine Learning (ML) algorithms have become front and center in our daily lives. 
 For example, ChatGPT alone was used  1.6 billion times globally in December 2023\footnote{https://explodingtopics.com/blog/chatgpt-users}. 
Another fast-growth research area is Time Series with applications such as human activity recognition using wearable devices \cite{singh:dami23}. 
Its basic form is a Univariate Time Series (UTS), which is a sequence of values recorded over time. Different time series can have different magnitudes, measure units, and semantics (for example speed recorded in km/h vs magnetic field recorded in Tesla).
A Multivariate Time Series (MTS)  is a set of sequences concurrently recorded over time, e.g., different ECG channels measuring a patient's heart activity \cite{UCRArchive2018}.

The popularity of ML algorithms is also accelerating another field, Explainable Artificial Intelligence (XAI): as the rate of automated decisions is increasing, it is necessary to explain the outcomes of these models. In the time series field, we can have a scenario where a user is executing a physical exercise using wearable devices: besides a reliable classifier, it is also desirable to have an explanation for the classification, so it is possible to give feedback to users about why the exercise was correctly or incorrectly executed.

\begin{figure}[t]
    \centering
    \subfloat[]{
        \includegraphics[width=0.35\textwidth]{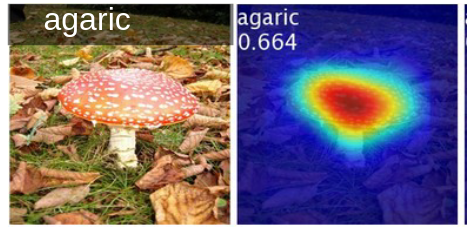} 
        }
    \subfloat[]{\includegraphics[width=0.55\textwidth]{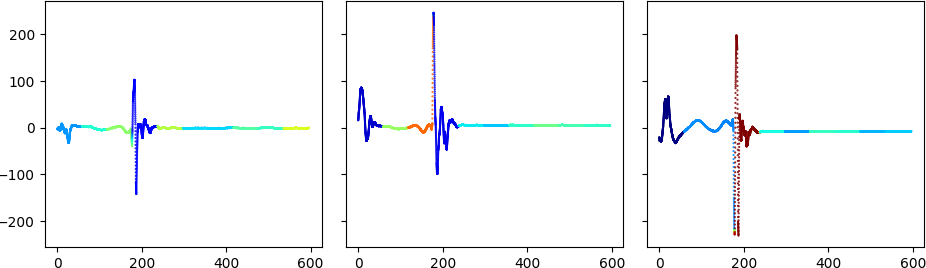}
        }
    \caption{ Example heat maps for two different input types: on the left, an image and the corresponding heat map (image from \cite{cam2015}), on the right, a three-channel time series from the Counter Movement Jump (CMJ) dataset \cite{le2019interpretable}: each of the 3 plots corresponds to one channel recording the $x,y,z$-axis acceleration (image from  \cite{serramazza2023evaluating}). In both cases, darker red means the more important the corresponding item is for the classification.}
    \label{fig:heat_maps_exmaples}
\end{figure}

In this paper we focus on Explanation Methods for Multivariate Time Series Classification (MTSC), in the form of saliency maps. A saliency map is a set of weights that represent the importance of each input item. For this reason, these XAI methods are also called attribution methods as they attribute importance to inputs.
Usually saliency maps are visualized through heat maps. Figure \ref{fig:heat_maps_exmaples} shows two different samples for Images and Time Series. The main idea is to use colors to intuitively represent input parts' importance, specifically the brighter the color (toward red), the more important the corresponding element was for the classification. 
\begin{figure}[t]
    \centering
    \includegraphics[width=0.9\textwidth]{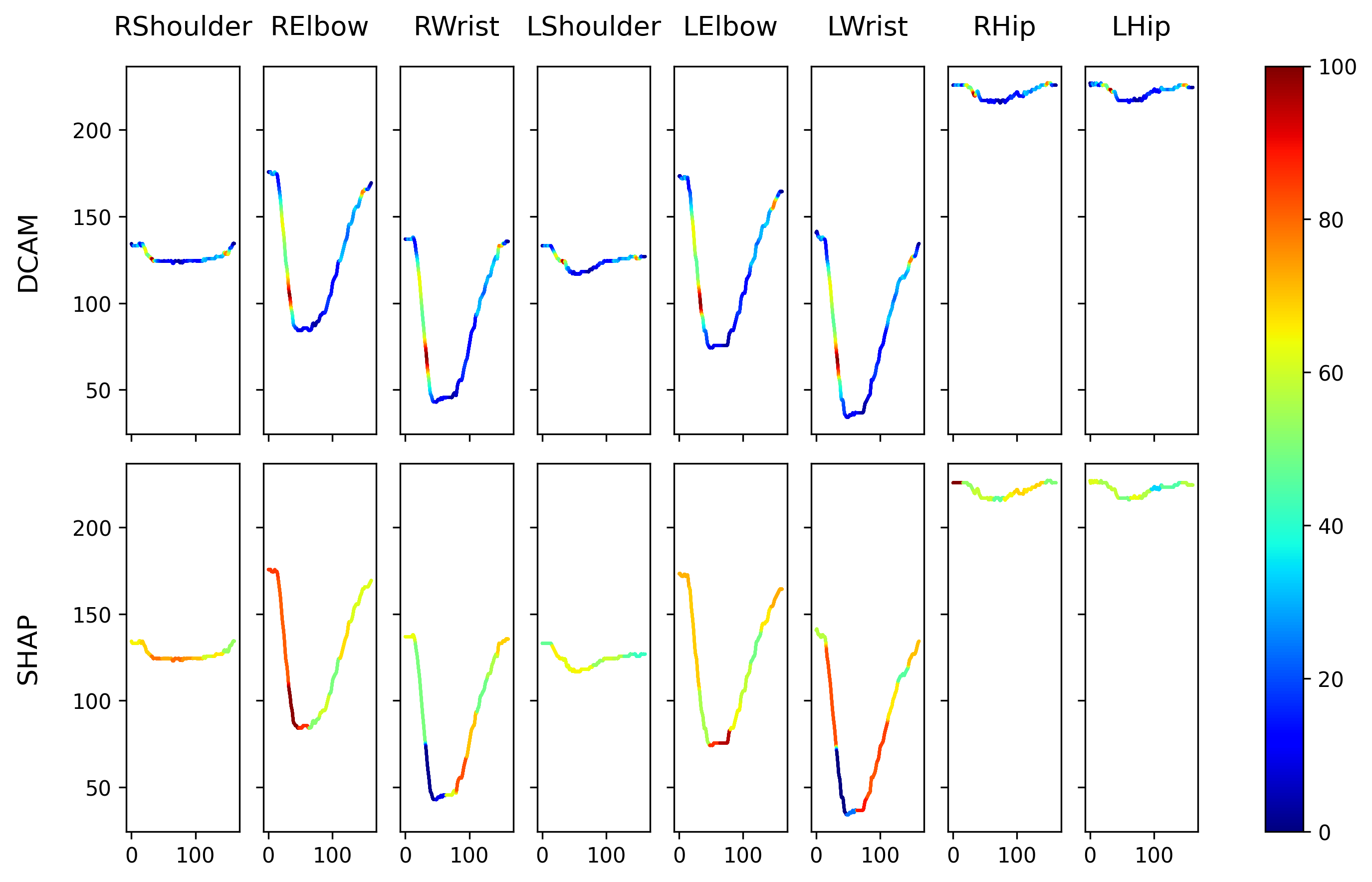} 
    \caption{ Same instance from the  Military Press (MP) dataset composed of 8 different channels recording $y$ positions of both Left and Right Wrists, Elbows, Shoulders and Hips. This figure visually shows a very frequent problem: two different attribution methods producing very different heat maps for the same instance  
    (image from \cite{serramazza2023evaluating}). }
    \label{fig:heat_maps_disagrement}
\end{figure}

An issue raised by XAI methods is what happens if two explanation methods highly disagree about attribution values for the same instance. Figure \ref{fig:heat_maps_disagrement} shows two heat maps providing an example of this frequent problem on the same instance of the Military Press exercise classification dataset \cite{singh:dami23}. 
Assessing the quality of different attribution methods is a difficult problem because, different from classification, ground truth for this task is often unavailable. Various evaluation metrics \cite{gomez2022metrics} have been proposed but we argue that they are limited in their capacity to tell us which explainer is better.
Other options in this case are either to use synthetic data where the ground truth is available or to rely on assessments based on data perturbations. For the latter, the input time series are first perturbed based on the saliency maps and fed to the models again. The drop in accuracy (before and after perturbation) is then measured: the higher the drop, the better the saliency maps \cite{nguyen2023amee,nguyen2024amee,turbe2023evaluation}. We argue that these methods are more reliable than other metrics as they quantify the quality of the explanation based on the related task (classification).

A saliency map based explanation is not only useful for relying on the classification, but also to make the explanations \textbf{actionable} by using them for instance to compute channel importance ranks and perform channel selection for MTSC.
Throughout our literature review we found very few attribution methods tailored to time series and most of them work only for specific model architectures, such as deep learning networks  \cite{boniol2022dcam,theissler2022,turbe2023evaluation}. Even fewer works carrying out evaluation among different attribution methods were conducted. 

Our main contributions in this paper are:
\begin{itemize}
\item We evaluate current attribution methods for MTSC, and analyze their strengths and limitations on both synthetic and real datasets.
\item We quantitatively compare multiple explanation methods for MTSC, starting from an existing methodology named InterpretTime. \textcolor{black}{We show that InterpretTime has some significant weaknesses: it requires train data augmentation which limits the type of classifiers that can be used. 
Additionally, it has poor efficiency in obtaining the explainer evaluation.}
\item We propose improvements to InterpretTime: \textcolor{black}{We do not use data augmentation and instead use multiple masks. We also show how grouping consecutive time points channel-wise helps reduce running time and improve the performance of explainers.}
\item We showcase the explainer actionability by using the best attribution methods for the task of channel selection in MTSC. All code and data is publicly available\footnote{
https://github.com/mlgig/xai4mtsc\_eval\_actionability
}. 

\end{itemize}
\section{Related Work}
\label{relwork}

In this section we review the state-of-the-art algorithms (SOTA) for MTSC and attribution methods, as well as methodologies to evaluate attributions. 

\subsection{Multivariate Time Series Classification}
\label{sec:classification}

Fortunately for the  MTSC task we have extensive empirical surveys pointing out SOTA algorithms \cite{IsmailFawaz2019,ruiz20mtsc}. Notable among them are the variations of ROCKET, a group of convolution-based and feature transformation classifiers. Features are extracted using randomly initialized convolutional kernels along with some key aspects such as Dilation or Percentage of Positive Values (PPV). Other variants including  MiniRocket \cite{dempster2021minirocket} and MultiRocket \cite{tan2022multirocket} are some of the best-performing algorithms for MTSC. On the other hand, deep learning-based (DL) classifiers are often adapted or take inspiration from image classification such as ResNet \cite{he2016deep} and InceptionTime \cite{ismail2020inceptiontime}. 
The key difference between these two fields is that images have only 3 channels while MTS has a varying number of channels. While DL algorithms are generally accurate, they behave poorly for datasets with a very high number of channels \cite{singh:dami23,ruiz20mtsc}.
ConvTran \cite{mao2022convtrans} is a recent transformer shown to be superior to other DL classifiers both in accuracy and noise robustness.

\subsection{Attribution Methods for Multivariate Time Series Classification}
\label{sec:attribution}

We can group attribution-based methods into mutually exclusive categories \cite{mao2022convtrans} such as \textit{Post-hoc} methods which produce a saliency map after the classification versus \textit{Self-explaining} methods where the classification and the explanation stages are merged, \textit{Model-specific} methods which can be applied only for specific classifiers versus  \textit{Model-agnostic} which can be applied to different methods. We can also divide them into \textbf{Gradient-based} methods that rely on the model gradient to explain the classification vs \textbf{Perturbation-based} approaches relying on perturbations of the inputs: the first ones are model-specific, most of the time used to explain Deep Learning classifiers, while the latter are generally model-agnostic. Table 1 shows a summary of popular attribution methods along with their characteristics.

Focusing on TS-specific methods some models have been proposed in the past such as \textit{MTEX-CNN} \cite{assaf2019mtex}, \textit{XTF-CNN} \cite{bi2021explainable} or \textit{CEFEs} \cite{maweu2021cefes} which are explainable Convolutional Neural Networks (CNN)s, \textit{TSViz} \cite{siddiqui2019tsviz} and \textit{TsXplain} \cite{munir2019tsxplain} which are tailored respectively to CNNs and NNs and the most recent \textit{dCAM} \cite{boniol2022dcam} which can be thought as a multidimensional version of CAM \cite{cam2015} tailored to MTSC. All of these methods share the same problem, they are model-specific which is undesirable, especially in a very heterogeneous field like MTSC where deep learning is not SOTA, as described before. One method that overcomes this limitation is the model-agnostic \textit{LimeSegment} \cite{sivill2022limesegment}, but it requires a large number of hyper-parameters to be specified.

\begin{table}[t]
    \centering
    \begin{tabular}{|c|c|c|c|c|}
        \hline
        Method & Specific/Agnostic & Gradient/Permutation & Tailored to TSC  \\
        \hline
        SHAP \cite{lundberg2017unified}  & Agnostic &  Perturbation & No \\ 
        dCAM \cite{boniol2022dcam} & Specific &  Gradient & Yes \\
        Integrated Gradient \cite{integratedGradient} & Specific &  Gradient & No \\
        DeepLift \cite{deepLift}  & Specific &  Gradient & No \\
        DeepLiftSHAP \cite{lundberg2017unified} & Specific &  Both & No \\
        GradientSHAP \cite{lundberg2017unified} & Specific &  Both & No \\
        KernelSHAP \cite{lundberg2017unified} & Agnostic &  Perturbation & No \\
        Feature Ablation \cite{fisher2019all} & Agnostic &  Perturbation & No \\
        Feature Permutation \cite{fisher2019all} & Agnostic &  Perturbation & No \\
        \hline
    \end{tabular}
    \caption{List of popular post-hoc attribution methods along with their characteristics.}
    \label{tab:attributions}
\end{table}

\subsection{Quantitative Evaluation of Attribution Methods for MTSC}

Quantitative evaluation of explanation methods includes quantitative metrics such as \textit{robustness, stability, sparsity} \cite{etmann2019connection,gomez2022metrics} and perturbation-based frameworks such as AMEE \cite{nguyen2023amee,nguyen2024amee} and InterpretTime \cite{turbe2023evaluation}.
\textit{Robustness} addresses the model's capacity to deal with noise while \textit{sparsity} measures how focused the saliency map is. Unfortunately, none of the above metrics give sufficient information on whether a particular saliency map is even better than a random one.
AMEE and InterpretTime present another approach where noise is inserted into the input time series at different levels and under the guidance of the saliency map. The drop in the accuracy of the classifiers (due to mislabelling the corrupted samples) indicates the quality of the explanation. While InterpretTime works for MTSC, AMEE is designed for UTSC but also works with MTS data by concatenating the channels \cite{serramazza2023evaluating}. \textcolor{black}{In this paper, we focus on the InterpretTime framework for MTSC and analyse this approach further.}

\subsection{Actionability of Explanation Methods}
\label{sec:actionability_literature_review}
We define explanation actionability as the possibility to use explanations for other computational tasks. One simple example of this is interpreting saliency maps as a way to approach feature importance in a classifier-independent way. The work of   \cite{clark2019does} found over-parametrizations in the BERT model by analyzing the self-attention layer. This can be considered as another actionability example, as this finding was used to improve the accuracy and efficiency of the model.  
Throughout a literature review we found only a few works showcasing explanation actionability in the TS domain. An example is  PUPAE \cite{der2024pupae} which uses counterfactual explanations to detect anomalies in time series in a domain-agnostic setting. Still, no improvement in a specific computational task was shown as a result of the PUPAE explanations.

\section{Background}

A multivariate time series can be presented as a $d \times L$ matrix $X$ where $d$ is the number of channels and $L$ is the length of the time series. Multivariate time series classification (MTSC) is the task of labeling unseen multivariate time series. The saliency map (or attribution) of $X$ is also a matrix $W$ of the same shape. The value $W_j^i$ at position $i,j$ ($j$-th item in $i$-th channel) indicates the importance of the corresponding input item ($X_j^i$) in the classification model. As the range of $W_j^i$ values can be different between methods, we normalize the saliency maps in the range $[0,1]$. 

\subsection{Classifiers}

In this paper we select state-of-the-art MTS classifiers MiniRocket, (d)ResNet, and ConvTran as the MTSC models to be explained. 
\label{sec:models}

\noindent\textbf{MiniRocket }\cite{dempster2021minirocket} is a convolution-based time series classifier that works with both univariate and multivariate TS. It is faster than its predecessor (Rocket) and successor (MultiRocket) while being comparably accurate. 
We used the implementation from \textit{tsai} \cite{tsai} since this PyTorch implementation is compatible with the Captum \cite{kokhlikyan2020captum} open source framework for model interpretability.\\
\textbf{ResNet} \cite{he2016deep} is a very popular CNN in the field of image classification which introduced the concept of skip/residual connection, i.e., to sum the input of a layer to its output. dResNet \cite{boniol2022dcam} is a variation of ResNet, requiring as input a $3D$ representation of the MTS, specifically each row is a specific circular shift of the original channel order. This change allows the network to learn the inter-relation of the channels. \\
\noindent\textbf{ConvTran} \cite{foumani2023improving} is a recent SOTA transformer that encodes positional information in multivariate time series data which relies on a new absolute and relative positional encoding designed specifically to deal with MTS data. On top of that it demonstrated much better noise resistance, differently from previous deep learning models and it is implemented in PyTorch and thus natively compatible with the Captum framework.
\subsection{XAI Methods}
\label{sec:xai_methods}

\textbf{Perturbation-based methods.} In this work we use Shapley Value Sampling to approximate SHAP value computation as proposed in \cite{lundberg2017unified}.
SHAP is considered one of the best attribution methods, extensively used in Image, Text and MTS fields, but it is computationally expensive \cite{turbe2023evaluation}, scaling poorly with the number of features.  Other permutation-based methods are Feature Ablation, Feature Permutation \cite{fisher2019all} which are very cheap to compute, and KernelSHAP which is a revisited version of Lime described in \cite{lundberg2017unified}. \\
\textbf{Gradient-based methods.} One other group of attribution methods is tailored to NNs, including methods such as Integrated Gradient \cite{integratedGradient}, DeepLift \cite{integratedGradient} or variations of them such as DeepLiftShap, or GradientSHAP proposed in \cite{lundberg2017unified}. \\
The last method we used in this work is dCAM \cite{boniol2022dcam} as representative of the model-specific methods tailored to TS previously described in Section \ref{sec:attribution}.
It can only be applied to modified versions of CNN having an average pooling layer, specifically, we applied it to the previously described dResNet. All attribution methods used in this paper, except for dCAM, are implemented using the Captum framework. 

\subsection{InterpretTime}

This is a framework for quantitative evaluation of attribution methods for MTSC using a metric named AUCS \cite{turbe2023evaluation}. The key idea is to use each attribution method for perturbing the test samples. The time series points are selected in order of the magnitude of their attribution value and are perturbed using Gaussian noise.

The quality of each saliency map is assessed by measuring the accuracy drop when the inputs $\{X_0 ,\dots ,X_N\}$ are perturbed according to the explanations $\{W_0, \dots ,W_N\}$ ($N$ is the test set length).  
More formally, a set of thresholds is defined $k=[0.05, 0.15, \dots , 0.95 , .1.0]$; for each $k$ two modified input versions are instantiated, $\bar{X}^{top}$ and $\bar{X}^{bottom}$ where respectively the top $k$-quantile and the bottom $(1-k)$ are replaced using values sampled from $\mathcal{N} (0,1) $; based on these modified inputs, the differences in models scores is computed as $\bar{S} = \frac{S(X) -S( \bar{X} )}  { S(X)}$ where $S$ is the model predicted score.  
Based on $\bar{S}^{top}$ and $\bar{S}^{bottom}$ the two metrics $AUC\bar{S}_{top}$ and $F1\bar{S}$ are defined, which respectively take into account the ability of a model to retrieve the most important items for a TS and the balance between retrieving the most important items and the least important ones.

InterpretTime makes some important assumptions: (1) training data is augmented with Gaussian noise, (2)  using $N(0,1)$ for perturbing the data is sufficient for ranking explanations, (3) only the positive attribution values are used for ranking explanations.

\subsection{Datasets}

\label{sec:datasets}

\textbf{Synthetic data.} The original dataset from \cite{turbe2023evaluation} is composed of 6 different channels each having 500 time points and containing different sine waves. The discriminative features come from 2 randomly selected channels,  in $100$ consecutive time points. In these positions, two other waves of higher frequency  $f_1$ and $f_2$ replace the original ones. Each of the other channels has a $0.5$ probability of having a non-discriminative \textit{square wave}. 
This binary classification task predicts whether or not $f_1 + f_2$ is above a given threshold.  
\textcolor{black}{In our implementation we extend the public code to add two non-discriminative channels (number $6$ and $7$)}; the test set is perfectly balanced having 500 positive and 500 negative instances, while the train set contains $7500$ samples with $3646$ positive ones.

\textbf{CMJ.} Counter-Movement Jump (CMJ) \cite{le2019interpretable} is accelerometer data collected from IMU sensors worn by ten participants while performing the counter-movement jump. The data has three channels which correspond to the X,Y,Z coordinates of the acceleration. Each sample is classified by the form of participants: acceptable form, legs bending during flight, and stumble upon landing. The original data is variable-length, as each jump is different. The shorter samples are padded with zeros to the length of the longest sample. The data is split (419 samples in the training set and 179 in the test set) so that there is no participant in both splits.

\textbf{MP.} Military Press (MP) \cite{singh:dami23} is a dataset recorded using video-capture of 56 participants doing physical exercises. Multivariate time series data are extracted from the videos by using OpenPose. The dataset has 50 channels tracking the X and Y coordinates of 25 body parts; each series length is 161. There are 1426 samples in the training set and 595 samples in the test set. Each sample is a repetition of the exercise by a participant, and no participant appears in both sets. The time series are manually labeled according to the form of the execution into: Normal (N), Asymmetrical (A), Reduced Range (R) and Arch (Arch). Domain experts suggest that the more important channels are elbows, wrists, shoulders, and hips in the $Y$ coordinates. In our experiments we used the previously listed 8 channels (left and right body parts),  centered by subtracting the mean from each of them as they have very different magnitude ranges.  

\textbf{ECG. }
This dataset was used for experiments in the InterpretTime paper \cite{turbe2023evaluation}. It has 12 channels, corresponding to 12 sensors that record electric signals from the heart. The task is to classify whether the signal carries a Right Bundle Branch Block (RBBB) or not. There are 5020 positive samples (RBBB present) and 1857 negative cases (RBBB absent). The dataset is further split into training (5846 samples) and test (1031 samples) sets; each sample has a length of 450.
\section{Methodology}

\subsection{MTS Chunking}
Many of the attribution methods are adapted directly from images. Since time series locality and closeness in temporal order could be important, we use time series chunking together with attribution methods. This has the potential to improve both the accuracy and the efficiency of the attribution methods. 
Using an instance of dimension $d \times L$, i.e., $d$ channels each of $L$ length, the classifier is trained using the standard data representation while some of the explanation methods, specifically the perturbation-based ones, can group different features into chunks computing only one attribution value for each group: the saliency map has still the same dimension as the original time series containing repeating attributions. From the perspective of the attribution method, this significantly reduces the number of features and can impact the speed and magnitude of the computed attribution. This is especially important for methods such as SHAP that are adversely affected by the time series length with regard to both runtime and magnitude of computed attributions.
To compare the quality of the produced saliency maps using this technique, we divided \textbf{each time-series channel} into $5$, $10$, $15$, and $20$ equal-sized chunks (in 4 different experiments) and aggregated the results into a single average score. We use the \textit{point-wise} and \textit{chunk-wise} terminologies to differentiate between the two scenarios. 
Finally we also tried to chunk MTS by grouping all channels together, ignoring the channel dimension, but this gave poor results. 

\subsection{Assessment based on Ground Truth}
\label{sec:gt_method}
For the synthetic data we used an assessment based on the ground truth; we describe here its steps.
The first one is to rescale the attribution matrices as they can have different magnitudes among different instances and different methods. To compare results with InterpretTime's original results, we took for each attribution $W$, the $max(0,W_j^i)$ (i.e., we consider only the positive values) and then normalize it in the range $[0,1]$. 
The second step is to instantiate an explanation ground truth matrix $\mathcal{G}$ for each instance in the test set and compare the explanation against it. This is a matrix where all items are set to $0$ except the ones corresponding to the \textit{discriminative elements} in the time series, having values set to $1$. In other words, this is the ideal saliency map that can be achieved from any attribution method. 
The last step is to compute scores, namely PR-AUC and ROC-AUC \cite{boniol2022dcam} comparing each saliency map against the corresponding $\mathcal{G}$.

\subsection{Improving InterpretTime}
\label{sec:interpret_time}
InterpretTime only uses values sampled from the $N(0,1)$  distribution to mask TS time points. We empirically verified that only relying on this distribution sometimes leads to a \textcolor{black}{flat} rank. 
This problem is due to the distribution shift from the train to the test set which in the original work was tackled by adding noise in the training procedure (more details in \cite{turbe2023evaluation}). We believe this is a downside because requiring a specific training procedure limits the classifiers that can be applied, e.g., we cannot use pre-trained classifiers \textcolor{black}{and it reduces the accuracy  of non deep learning classifiers.}
We instead \textcolor{black}{\textbf{do not augment the training data as done in \cite{turbe2023evaluation}}} and fixed this problem by using \textbf{multiple masks}. Besides $N(0,1)$, we use Local and Global Gaussian and Local and Global mean as done in AMEE \cite{nguyen2023amee} and the \textit{zeros} distribution. Substituting the element $x_t^c$ (i.e., channel $c$, $t$-th time point of one element $x$ of the test set $X$), the replacing value $r_t^c$ is $r_t^c=0$ for the zeroes distribution, while the others depend on the following means and standard deviations:

\begin{equation*}
    \mu_t^c  = \frac{1}{N} \sum_{x_t^c \in X}  x_t^c  \hspace{20pt} 
    \sigma_t^c = \frac{1}{N-1} \sum_{ x_t^c \in X}  (x_t^c - \mu_t^c;)^2 
\end{equation*}
\begin{equation*}
    \mu=\frac{1}{N} \sum_{x \in X} \sum_{c=1} ^{d} \sum_{t=1} ^{L}   x_t^c   \hspace{14pt} 
     \sigma = \frac{1}{N-1} \sum_{ x \in X} \sum_{c=1} ^{d} \sum_{t=1} ^{L}  (x_t^c - \mu_t^c)^2;
\end{equation*}
Specifically for local mean $r_t^c=\mu_t^c$, for local Gaussian $r_t^c \sim \mathcal{N}(\mu_t^c ,\sigma_t^c)$, for global mean $r_t^c=\mu$ and for global Gaussian $r_t^c \sim \mathcal{N}(\mu, \sigma)$. \\

We first run InterpretTime using all the previously described masks recording the score of each attribution method for each mask. We then discard the masks where $AUC\tilde{S}_{top}$ of that attribution method is not greater than the $AUC\tilde{S}_{top}$ of the random attribution by at least $15\%$ (we empirically found this threshold to separate between meaningless and meaningful). Results for all non-discarded masks are then averaged to get a final score for each attribution method. 

\subsection{Actionable XAI: Channel Selection for MTSC using Attribution}

\label{sec:channel_selection}

Using one attribution method, we computed a saliency map $W$, having the same dimension as the input, i.e., $d \times L$, for each instance in the dataset $D$. We then average among all saliency maps and among the time points dimensions to compute the $d$-dimensional vector $r$.
\begin{equation*}
    r= \frac{1}{|D|} \frac{1}{L} \sum_{ W \in D}  \sum_{i=1}^L W_i^c
\end{equation*}
Each component of this vector represents a raw importance score of the corresponding channel in the dataset.
We can use these scores to rank channels and for channel selection.

\section{Experiments}

All classifiers described in Section \ref{sec:models} were trained 5 different times using all datasets described in Section \ref{sec:datasets}. The accuracy for the most accurate models are shown in Table \ref{tab:accuracy}. For MiniRocket, we use default parameters and $5$-fold cross-validation for  Logistic Regression. For (d)ResNet we allowed up to 300 epochs using an early stopping after 50 consecutive epochs of non-improving accuracy. For ConvTran, we use the procedure of the original authors, namely to take the best model (in our case best accuracy) found in 100 epochs. Finally, as opposed to the approach taken in InterpretTime \cite{turbe2023evaluation} we did not use any data augmentation in the training procedure. We add a random saliency map as a baseline,  obtained by sampling attribution weights uniformly at random in $[0,1]$.

\begin{table}[t]
    \centering
    \begin{tabular}{|c|c|c|c|c|c|c|c|}
        \hline
        Model/Dataset & Synthetic & CMJ    & MP domain expert channels   & MP all channels & ECG \\
        \hline
        ResNet      &       0.868 & \textbf{0.966}  & 0.786 & 0.615  & 0.955  \\
        dResNet     &       0.932 & 0.944  & 0.781 & 0.504 & 0.931  \\
        MiniRocket  &       0.929 & 0.949 & 0.757 & 0.805  & 0.961  \\
        ConvTran    &        \textbf{0.967} & 0.944  & \textbf{0.844} & \textbf{0.831}  & \textbf{0.967}   \\
        Tabular-baseline & 0.519     & 0.743 & 0.571 & 0.593 & 0.949  \\
        \hline
    \end{tabular}
    \caption{Classifier accuracy for all datasets. Tabular-baseline is the best accuracy achieved between RidgeCV and LogisticRegressionCV.}
    \label{tab:accuracy}
\end{table}
All experiments were run using a machine having one AMD EPYC 9654P CPU (96 cores, 192 threads), NVIDIA GeForce RTX 4090 GPU, 1.5Tb of RAM.
\subsection{Validating InterpretTime Results} 
\label{sec:validating}

\begin{table}[t]
    \begin{tabular}{cc}
    
        \begin{minipage}{.55\linewidth} 
                \begin{tabular}{|c|c|c|c|c|}
                    
                    \hline
                    Model  &  XAI method   & AP   & ROC & Time \\
                    \hline
                    
                    ConvTran  &  \textbf{F. Ablation}  & \textbf{.370} & \textbf{.660} & 12m \\ 
                    ConvTran  &  F. Permutation  & .306 & .656 & 15m\\ 
                    ConvTran  &  Int. Gradients  & .270 & .611 & 15.5s \\ 
                    ConvTran  &  DeepLift       & .242 & .611 & .85s \\ 
                    ConvTran  &  DeepLiftShap  & .233 & .615 & 2.3s \\ 
                    ConvTran  &  Saliency  & .197 & .592 & .38s \\ 
                    ConvTran  &  SHAP       & .194 & .590 & 6.3h \\ 
                    ConvTran  &  GradientShap  & .10 & .548 & 1.65s \\ 
                    ConvTran  &  KernelShap  & .052 & .500 & 4m \\ 
                    
                    \hline
                    MiniRocket  & \textbf{F. Ablation}  & \textbf{.186} & .616 & 42m\\ 
                    MiniRocket  &  F. Permutation  & .182 & \textbf{.659} & 43m \\ 
                    MiniRocket  &  SHAP  & .088 & .554 & 17h \\ 
                    MiniRocket  &  KernelShap  & .052 & .501 & 4m \\ 
                    \hline
                    NA  & random & .052 & .500 & \\ 
                    \hline
                \end{tabular}
        \end{minipage} &

        \begin{minipage}{.55\linewidth}
                \begin{tabular}{|c|c|c|c|c|}
                    
                    \hline
                    Model  &  XAI method   & AP   & ROC & time\\
                    
                    \hline
                    ResNet  &  \textbf{SHAP}  & \textbf{.175} & .634 &  27m \\ 
                    ResNet  &  F. Ablation  & .168 & .618 & 58s\\ 
                    ResNet  &  Int. Gradients  & .153 & .608 & 3s \\ 
                    ResNet  &  DeepLift  & .149 & .603 & .3s \\ 
                    ResNet  &  F. Permutation  & .13 & \textbf{.638} & 64s\\ 
                    ResNet  &  DeepLiftShap  & .102 & .595 & 7s \\ 
                    ResNet  &  Saliency  & .067 & .542 & .15\\ 
                    ResNet  &  GradientShap  & .074 & .547 & .35 \\ 
                    ResNet  &  KernelShap  & .052 & .502 & 1.9m \\ 
                    
                    \hline
                    dResNet  &  \textbf{dCAM}  & \textbf{.396} & \textbf{.914} & 29m\\ 
                    dResNet  &  F. Ablation  & .154 & .613 & 1.9h \\ 
                    dResNet  &  SHAP  & .131 & .612 & 53h \\ 
                    dResNet  &  F. Permutation  & .127 & .637 & 1.9h \\ 
                    dResNet  &  KernelShap  & .051 & .501 & 0.2h \\
                    \hline
                \end{tabular}
        \end{minipage} 
    \end{tabular}
\caption{\textbf{Synthetic data.} AP and ROC for different attribution methods; results are sorted by AP and grouped by different classifiers. For ConvTran and ResNet we tested both the gradient and the perturbation-based explanations, while for MiniRocket and dResNet we can use only the perturbation-based methods since they use non-differentiable operations; dCAM was custom developed for dResNet.}
\label{tab:validation_GT}
\end{table}

\begin{table}[htb]
    \begin{tabular}{cc}
        \begin{minipage}{.55\linewidth}
            \begin{tabular}{ll}
                \begin{tabular}{|c|c|c|c|}
                    
                    \hline
                    Model  &  XAI method   & \ $AUC\tilde{S}_{top}$    & $F1\mathcal{S}$  \\
                    
                    \hline
                    ConvTran & \textbf{F. Ablation} & \textbf{0.619} & \textbf{0.256} \\ 
                    ConvTran & F. Permutation & 0.505 & 0.241\\ 
                    ConvTran & SHAP & 0.468 & 0.211\\ 
                    ConvTran & DeepLift & 0.460 & 0.165\\ 
                    ConvTran & Int. Gradients & 0.460 & 0.173\\ 
                    ConvTran & DeepLiftShap & 0.449 & 0.168\\ 
                    ConvTran & Saliency & 0.439 & 0.256\\ 
                    ConvTran & GradientShap & 0.435 & 0.175\\ 
                    ConvTran & KernelShap & 0.429 & 0.165\\ 
                    \hline
                    ConvTran & Random & 0.433 & 0.166\\ 
                    \hline
                    
                    \hline
                    MiniRocket & \textbf{F. Ablation} & \textbf{0.515} & \textbf{0.224}\\ 
                    MiniRocket & F. Permutation & 0.463 & 0.218\\ 
                    MiniRocket & SHAP & 0.421 & 0.177\\ 
                    MiniRocket & KernelShap & 0.400 & 0.151\\ 
                    \hline
                    MiniRocket & Random & 0.402 & 0.156\\
                    \hline 
                    
                \end{tabular}
            \end{tabular}
        \end{minipage} &
    
        \begin{minipage}{.55\linewidth}
            \begin{tabular}{ll}
                \begin{tabular}{|c|c|c|c|}
                    \hline
                    Model  &  XAI method   &  \ $AUC\tilde{S}_{top}$   &  $F1\mathcal{S}$  \\
                    \hline
                    ResNet & \textbf{F. Ablation} & \textbf{0.607} & \textbf{0.238}\\ 
                    ResNet & Int. Gradients & 0.586 & 0.218\\ 
                    ResNet & DeepLift & 0.571 & 0.215\\ 
                    ResNet & F. Permutation & 0.543 & 0.249\\ 
                    ResNet & DeepLiftShap & 0.515 & 0.218\\ 
                    ResNet & SHAP & 0.466 & 0.192\\ 
                    ResNet & GradientShap & 0.429 & 0.179\\ 
                    ResNet & Saliency & 0.341 & 0.233\\ 
                    ResNet & KernelShap & 0.328 & 0.130\\ 
                    \hline
                    ResNet & Random & 0.321 & 0.121\\ 
                    \hline
                
                    \hline
                    dResNet & \textbf{F. Ablation} & \textbf{0.667} & \textbf{0.259} \\ 
                    dResNet & F. Permutation & 0.511 & 0.239 \\ 
                    dResNet & SHAP & 0.416 & 0.169 \\ 
                    dResNet & dCAM & 0.393 & 0.273 \\ 
                    dResNet & KernelShap & 0.381 & 0.150 \\ 
                    \hline
                    dResNet & Random & 0.391 & 0.150 \\
                    \hline
                
                \end{tabular}
            \end{tabular}    
        \end{minipage} 
    \end{tabular}
    
\caption{\textbf{Synthetic data.} Average  $AUC\tilde{S}_{top}$ and $F1\mathcal{S}$ for each method (including Random) using \textbf{multiple masks}: global mean, local mean, and zeros distribution. The only time the standard deviation among these 3 masks was higher than 0.01 was for FeaturePermutation using dResNet. The runtime is the same as for Table \ref{tab:validation_GT}.}
\label{tab:InterpretTime_moreMasks}
\end{table}
The first experiment was to \textcolor{black}{compare to} the results presented in \cite{turbe2023evaluation} using synthetic data.
The InterpretTime methodology assesses the quality of saliency maps only by perturbing the time series, and the original study does not present a comparison to ground truth explanations for the synthetic data. We compared the explanation methods using the ground truth methodology illustrated in Section \ref{sec:gt_method} (results are reported in Table \ref{tab:validation_GT}). \\
Focusing on the XAI methods also tested in \cite{turbe2023evaluation}, we also found that SHAP is generally the best method (in our case for 3 out of 4 classifiers), KernelShap achieved all the time results very close to the random method and all gradient-based methods are in the middle between these two extremes. Different  from the original work, we also found that SHAP has some of the poorest metrics using ConvTran, moreover GradientShap which is a competitive method in the original work, in our study is the second to last in both tested cases. 
In our experiments, we have also added FeatureAblation and FeaturePermutation, which can be seen as very simple and fast alternatives to SHAP. A final remark is that metrics are generally poor, suggesting that there is a weak overlap between the assessed saliency maps and the ground truth $\mathcal{G}$.

In our second experiment, we applied InterpretTime using only the $N(0,1)$ mask to compute the $AUC\tilde{S}_{top}$ score for each attribution, to directly compare with the original paper and with the metrics obtained using the ground truth.
\textcolor{black}{Under the setup we tested, namely no train data augmentation, we were not able to reproduce the results from the original paper.} Moreover, the ranking obtained using $AUC\tilde{S}_{top}$ is almost flat and very different than the one obtained using the metrics computed with ground truth. We show detailed results in the Appendix.
In Section \ref{sec:improving} we discuss what are the main sources of the problems pointed out by these two experiments (flat  InterpretTime ranking and poor metrics for both cases) and the solutions we found for them.

\subsection{Improved XAI Evaluation Methodology}
\label{sec:improving}

\subsubsection{Using multiple masks.}
Using only the $N(0,1)$ distribution, as the time series are progressively corrupted, all of the 4 classifiers predict the same class for every instance, specifically the negative one for synthetic data \textcolor{black}{(see Appendix).}
Using this dataset, the $N(0,1)$ distribution, and the Local and Global Gaussian distributions are below the $15\%$ threshold, that is why they were excluded from scores computation, while the Global and Local mean and zeros distributions are above, showing very close results. Table \ref{tab:InterpretTime_moreMasks} reports the final metrics.  
We notice that for the best methods $AUC\tilde{S}_{top}$ is now remarkably higher than the corresponding random method, thus using the synthetic dataset our modified InterpretTime version leads to effective ranking, where metrics differences among different methods are now evident, and also have a better alignment with the ground truth results, especially for the ResNet classifier. In Section \ref{sec:real_world_data} another ranking problem due to the usage of $N(0,1)$ is shown.
A final remark is that the $F1\mathcal{S}$ scores are most of the time proportional to $AUC\tilde{S}_{top}$: from now on we mostly focus on the $AUC\tilde{S}_{top}$ score in our analysis.

\begin{table}[t]
    \centering
    \begin{tabular}{|c|c|c|c|c|c|c|c|c|c|}
        \hline 
        Classifier & Method & Avg Time & AP & ROC & $AUC\tilde{S}_{top}$   & $F1\mathcal{S}$ \\
         \hline
        ConvTran & \textbf{F. Ablation}   & 14s      & \textbf{0.464} & 0.609 & 0.594 & 0.184   \\ 
         ConvTran & SHAP   & 7.30m           & 0.457 & 0.602 & \textbf{0.597} & 0.186 \\
        ConvTran & F. Permutation   & 18s  & 0.438 & \textbf{0.646} & 0.548 & \textbf{0.217} \\        
        ConvTran & KernelShap   & 2.9m      & 0.161 & 0.556 & 0.452 & 0.174 \\ 
        
        ConvTran & random & / & 0.052 & 0.501 & 0.26 & 0.116 \\ 
         \hline 
         MiniRocket & \textbf{SHAP}
         & 20.2m         & \textbf{0.477} & \textbf{0.747} & \textbf{0.674} & 0.228 \\ 
        MiniRocket & F. Ablation  & 51s    & 0.465 & 0.738 & 0.628 & \textbf{0.229} \\ 
        MiniRocket & F. Permutation   & 51s  & 0.403 & 0.743 & 0.497 & 0.16 \\ 
        MiniRocket & KernelShap   & 2.2m    & 0.131 & 0.567 & 0.415 & 0.146 \\
        
        MiniRocket & random & / & 0.052 & 0.501 & 0.255 & 0.89 \\ 
          \hline 
        ResNet & \textbf{SHAP}   & 31.6s             & \textbf{0.544} & \textbf{0.761} & \textbf{0.625} & \textbf{0.195} \\ 
        ResNet & F. Ablation   & 1.22s      & 0.401 & 0.724 & 0.582 & 0.191 \\ 
        ResNet & F. Permutation   & 0.68s  & 0.356 & 0.726 & 0.481 & 0.200 \\ 
        ResNet & KernelShap   & 33s         & 0.154 & 0.594 & 0.351 & 0.141 \\
        
        ResNet & random & / & 0.052 & 0.501 & 0.172 & 0.087 \\ 
          \hline 
        dResNet & \textbf{SHAP}   & 7.30m            & \textbf{0.584} & 0.747 & 0.657 & 0.208 \\ 
        dResNet & F. Ablation   & 17s       & 0.523 & \textbf{0.749} & \textbf{0.666} & \textbf{0.214} \\ 
        dResNet & F. Permutation   & 18s    & 0.458 & 0.747 & 0.591 & 0.209 \\ 
        dResNet & KernelShap   & 33s        & 0.227 & 0.631 & 0.566 & 0.196 \\ 
        
        dResNet & random &   / & 0.052 & 0.501 & 0.172 & 0.067 \\ 
        \hline
    \end{tabular}
    \caption{\textbf{Synthetic data.} Average scores for perturbation-based methods across \textbf{different chunk sizes}. For each method, the computation time reduces significantly, while metrics increase.} 
    \label{tab:improving_GT_chunks}
\end{table}

\subsubsection{Better alignment with ground truth. }
The other problem pointed out in Section \ref{sec:validating} was that scores, especially AP and ROC achieved from the Ground Truth experiment are generally poor: we identified the problem is due to the high number of attribution values to be computed since every instance of the synthetic dataset has $4000$ different time points (Section \ref{sec:datasets}). We address this by using chunking to reduce the number of features considered by the attribution method.
In Table \ref{tab:improving_GT_chunks} we report scores for \textbf{chunk-wise} experiments obtained by averaging among different chunk sizes (in addition to different masks). Other than a big drop in the running time (as compared to point-wise runtime results in Table \ref{tab:validation_GT}), it is possible to appreciate that the metrics also improved (AP and ROC all the times, $AUC\tilde{S}_{top}$   and $F1\mathcal{S}$ most of the time). We provide more results in the Appendix.  
Comparing point-wise and chunk-wise results it is clear that the method benefiting the most in this scenario is SHAP which is 3 out of 4 times the best model.
\begin{table}[ht!]

    \begin{tabular}{|c|c|c|c|c|c|c|c|c|c|}
        \hline
         \multicolumn{2}{|c|}{} & \multicolumn{4}{c|}{CMJ} & \multicolumn{4}{c|}{MP} \\
        \hline 
          \multicolumn{2}{|c|}{} & \multicolumn{2}{c|}{Point-wise} & \multicolumn{2}{c|}{Chunks avg.} &\multicolumn{2}{c|}{Point-wise} & \multicolumn{2}{c|}{Chunks avg.}  \\
        \hline
        Model  &  XAI method   & $AUC\Tilde{{S}}$   & Time & $AUC\Tilde{{S}}$   & Time    & $AUC\Tilde{{S}}$   & Time & $AUC\Tilde{{S}}$   & Time  \\
        
         \hline
        ConvT. & DeepLift & 0.667 & .15s & / & /  & 0.671 & .89s & / & / \\ 
        ConvT. & DeepLiftShap & 0.66 & .24s & / & /  & 0.672 & .83s & / & / \\ 
        ConvT. & F. Ablation & 0.615 & 10.7s & 0.677 & .63s   & 0.648 & 19.4s & 0.701 & 1.35s \\ 
        ConvT. & F. Permutation & 0.585 & 13.1s & 0.659 & .74s   & 0.597 & 64.2s & 0.638 & 2.11s \\ 
        ConvT. & GradientShap & 0.679 & .17s & / & /  & 0.691 & .57s & / & / \\ 
        ConvT. & Int. Gradients & 0.683 & 1.3s & / & /  & 0.699 & 2.38s & / & / \\ 
        ConvT. & KernelShap & 0.536 & 9.4s & 0.61 & 31s   & 0.486 & 1.95m & 0.599 & 1.76m \\ 
       ConvT. & Saliency & 0.554 & .5s & / & /  & 0.433 & .47s & / & / \\ 
        ConvT. & SHAP & \textbf{0.777} & 5.3m & 0.741 & 17s   & \textbf{.754} & 26.8m & 0.749 & 1.65m \\ 
        \hline
        ConvT. & Random & 0.5 & / &.482& /  & 0.402 & / &.472 & / \\ 
                                             
        \hline
        MiniR. & F. Ablation & 0.448 & 61.4s & 0.519 & 1.4s  & 0.535 & 2.2m & 0.547 & 8.49s \\ 
        MiniR. & F. Permutation & 0.464 &   62s & 0.499 & 1.4s   & 0.508 & 2.3m & 0.53 & 8.53s \\ 
        MiniR. & KernelShap & 0.469 & 29s & 0.495 & 23s   & 0.469 & 1.3m & 0.483 & 1.72m \\ 
        MiniR. & SHAP & \textbf{0.632} & 25m & 0.621 & 32s   & \textbf{.663} & 54m    & 0.617 & 3.7m \\ 
        \hline
         MiniR. & Random & 0.433 & / &.417& /  & 0.432 & / &.396& / \\
        \hline

        ResN. & DeepLift & 0.565 &  .16s & / & /   & 0.562 & .197s & / & / \\ 
        ResN. & DeepLiftShap & 0.572 & .51s & / & /  & 0.54 & 3.1s & / & / \\ 
         ResN. & F. Ablation & 0.531 & 8.1s & 0.577 & .22s   & 0.534 & 11.6s & 0.5 & .805s \\ 
        ResN. & F. Permutation & 0.444 & 9.8s & 0.401 & 21s   & 0.52 & 13s & 0.543 & .834s \\ 
        ResN. & GradientShap & 0.575 & .17s & / & /  & 0.529 & .194s & / & / \\ 
        ResN. & Int. Gradients & 0.574 & 1.1s & / & /                      & 0.574 & 1.48s & / & / \\
        ResN. & KernelShap & 0.42 & 11s & 0.431 & 5.5s   & 0.485 & 36s & 0.533 & 19.07s \\ 
        ResN. & Saliency & 0.261 & 11s & / & /  & 0.441 & .067s & / & / \\ 
        ResN. & SHAP & \textbf{.648} & 4m & 0.634 & 5.1s & \ 0.583 & 5.3m & \textbf{0.641} & 19.6s \\ 
        \hline
        ResN. & Random & 0.365 & / &.212& / & 0.44 & / &.443 & / \\ 
        \hline
        dResN. & dCAM & 0.469 & 4.9s & / &/  & 0.358 & 4.38m & / & / \\ 
        dResN. & F. Ablation & 0.459 & 23s & 0.453 & .31s   & \textbf{.455} & 6.5m & 0.368 & 3.4s \\ 
        dResN. & F. Permutation & 0.429 & 28s & 0.418 & .29s   & 0.42 & 6.7m & 0.393 & 3.2s \\  
        dResN. & KernelShap & 0.428 & 19s & 0.471 & 6s   & 0.403 & 2.46m & 0.437 & 18.9s \\ 
        dResN. & SHAP & .559 & 12m & \textbf{0.587} & 7.1s   & 0.4 & 1.92h & 0.519 & 1.18m  \\ 
        \hline
        dResN. & Random & 0.385 & / & 0.360 & / & 0.349 &  / & 0.359 & / \\
        \hline

    \end{tabular}

\caption{\textbf{CMJ and MP datasets}. $AUC\Tilde{{S}}$ stands for $AUC\tilde{S}_{top}$. Excluded masks for CMJ are Global Gaussian for all classifiers both point and chunk-wise, Local Gaussian is included only for ConvTran and ResNet chunk-wise. The only excluded mask for MP is Global Gaussian.}
\label{tab:real_world_datasets}
\end{table}

\subsection{Real World Data}
\label{sec:real_world_data}
In Table \ref{tab:real_world_datasets} we reported the results for the real-world CMJ and MP datasets, while ECG results are in the Appendix. Table \ref{tab:accuracy} shows that on the ECG dataset the  \textit{tabular baseline classifier} has accuracy close to the MTS SOTA classifiers.  This means that the temporal information in this dataset is not important for the classification. We include the results for this dataset in the appendix to have another comparison with \cite{turbe2023evaluation}, but we think this problem is too easy to benchmark attribution methods for MTSC. \\
\textbf{CMJ.} Every time the best-performing method is SHAP point-wise, the second to best is SHAP chunk-wise having a small performance degradation metric-wise and gaining a lot in computation speed compared to the first one. Most of the time other perturbation-based explanations have better metrics when used with the chunking strategy (except for dResNet). For instance FeatureAblation chunk-wise is competitive compared to the best gradient-based methods.  \\
\textbf{MP.} Results on MP are similar to the CMJ ones. In this case, Feature Ablation chunk-wise is the second-best method using ConvTran. The ResNet classifier is the only one where performance does not improve using the chunk-wise strategy.  
Using this dataset we also found another problem related to only using $N(0,1)$ for perturbing time series: most of the time 2 different masks provide 2 different best attributions, even though the margin between these scores and the corresponding random ones is by far larger than the previous threshold we set up. In the appendix we provided more results for the worst case, i.e., ConvTran with $3$ different masks provides $3$ different best attribution methods. This suggests again that the distribution used to perturb time-series can have a huge impact on results and our strategy of averaging up to $6$ different masks, unlike using only $N(0,1)$, can mitigate this problem.  \\
Taking into account all results from the 3 datasets suggests that the simpler the dataset (CMJ) the more competitive the gradient-based methods are. The harder the dataset (MP and synthetic data accuracy, see Table \ref{tab:accuracy}) the more effective the perturbation-based ones are, especially when used chunk-wise. 

\subsection{Actionability}
We demonstrate the actionability of the saliency maps with the task of channel selection on the \textbf{synthetic} and \textbf{MP} dataset. 
On the \textbf{synthetic} dataset, the two added channels are the least important ones (based on the ground truth). All methods but KernelShap are able to identify these two channels as their scores are the lowest. This agrees with our previous finding in which KernelShap is one of the poorer explainers (Table \ref{tab:InterpretTime_moreMasks} and \ref{tab:improving_GT_chunks}).
\begin{table}[t]
    \centering
    \begin{tabular}{ |c|c|c|c|c|c|c|c| }
       \hline
       Selection    &  n.channels &dResNet  & ResNet & ConvTran         & MiniRocket        & Rocket  \\
       \hline
       XAI-based & 8  &    0.771  &  0.746 &  0.828           &   \textbf{0.81}            &   \textbf{0.855} \\   
       Domain expert & 8  &  \textbf{0.781}   &  \textbf{0.771} &   \textbf{0.844}  &  0.769   &   0.831 \\
        Elbow Cut Pairwise \cite{dhariyal2023scalable} & 14 & 0.573 & 0.650 & 0.808 & 0.791 & 0.836 \\
        All available & 50 &   0.504   &  0.615 &  0.831           & 0.805             &   0.823 \\
        \hline
    \end{tabular}
    \caption{Accuracy for different MP channel selection approaches. }
    \label{tab:actionability}
\end{table}

A second experiment was conducted using the original \textbf{MP} dataset having 50 channels. The eight most important channels selected by the domain expert are: right and left Shoulder, Elbow, Wrist and Hip using the $Y$ coordinates. ECP \cite{dhariyal2021fast}, a state-of-the-art channel selection method for MTSC, selects 14 channels. For our proposed method, explained in Section \ref{sec:channel_selection}, we used ConvTran as the backbone model and explained it with SHAP and Feature Ablation in the 10 chunks fashion. Since SHAP and FeatureAblation agree on the 8 most important channels (right Wrist and Left Knee $Y$ channels, Right Ankle and Heel $X$ channels, and all 4 channels for Elbow), but slightly disagree about their order, we can consider the total time to compute channel ranking as the Feature Ablation running time which was approximately one minute. Finally we evaluate the performance of different channel selection methods by re-training and testing on the selected channels with different classifiers. As done before, all classifiers were trained 5 times, only the best accuracy was reported (Table \ref{tab:actionability}). Overall,  channel selection benefits all classifiers in this dataset. XAI-based and domain expert selection are the best, with comparable accuracy across the classifiers. Our XAI-based selection achieved the highest accuracy with MiniRocket and ROCKET, while domain-expert selection had the highest accuracy for the rest. 
\section{Conclusion}

In this work we have \textcolor{black}{investigated} the recent InterpretTime methodology to compare different attribution methods for MTSC, using an assessment based on ground truth which was not considered in the original study. We found that, \textcolor{black}{if we remove the train data augmentation used in the original study}, for some dataset/classifier pairs the methodology cannot be run using \textcolor{black}{only the $N(0,1)$ distribution mask as it provides a flat rank}; we \textcolor{black}{mitigated} this problem by introducing multiple masks and selecting only the ones producing a reasonable margin between the best and the random AUC score. 
We also improved the AUC scores of attribution methods, while reducing running time, by using chunking, i.e., grouping consecutive features in each channel. This is possible only for perturbation-based models as they can mask features together. 
Overall, SHAP is the best method with regard to accuracy. 
The simple FeatureAblation method is surprisingly not far from SHAP accuracy, but is much faster and overall better than other perturbation-based methods. We recommend using perturbation-based methods as the gradient-based methods are model-specific (deep learning) and cannot take into account the MTS locality as done with the chunking strategy. 
Finally, we run experiments focused on actionability, showing that the best attribution methods can correctly identify the most important channels in a dataset, using both synthetic and real-world datasets.

\textbf{Future Work.} 
In this work, we showed that attribution methods can be made actionable, specifically we used them for the channel selection task. More work can be done to improve the ideas presented in this paper (e.g., determine the number of channels to be selected), and other tasks can be approached using similar ideas, for example using XAI for feature engineering or model optimization.

More work is also required to better understand and address the mask-selection problem highlighted in the InterpretTime method. 
\textcolor{black}{Selecting one single distribution for masking 
is not enough, because different masks can lead to different ranking.}
In this work we mitigated this problem by averaging up to 6 different masks but this point is still an open problem that requires more investigation. 

\section*{Acknowledgments}
\textcolor{black}{
We thank the anonymous reviewers for their constructive feedback. We are grateful to Jiawen Wei and Gianmarco Mengaldo for detailed discussions on the original InterpreTime methodology. 
We thank all researchers working on time series and explainable AI who have made their data, code
and results open source to help the reproducibility of research methods in this area. This work was funded by Science Foundation Ireland through the SFI Centre for Research Training in Machine Learning (18/CRT/6183) and the Insight Centre for Data Analytics (12/RC/2289\_P2). For the purpose of Open Access, the author has applied a CC BY public copyright licence to any Author Accepted Manuscript version arising from this submission.
}

\newpage
\newcommand{\subdir}{supplementary}
\section*{Appendix}

\subsection*{InterpretTime synthetic data $N(0,1)$ mask results}

\begin{table}[h]
    \begin{tabular}{cc}
        \begin{minipage}{.55\linewidth}
                \begin{tabular}{|c|c|c|c|}
                    
                    \hline
                    Model  &  XAI method   & $AUC\tilde{S}_{top}$   & $F1\mathcal{S}$ \\
                    
                    \hline
                    ConvTran & \textbf{F. Ablation} & \textbf{0.463} & 0.205 \\ 
                    ConvTran & F. Permutation & 0.459 & \textbf{0.220} \\ 
                    ConvTran & SHAP & 0.448 & 0.200 \\ 
                    ConvTran & Int. Gradients & 0.444 & 0.166 \\ 
                    ConvTran & DeepLift & 0.445 & 0.157 \\ 
                    ConvTran & DeepLiftShap & 0.445 & 0.165 \\ 
                    ConvTran & Saliency & 0.444 & 0.169 \\ 
                    ConvTran & GradientShap & 0.443 & 0.178 \\ 
                    ConvTran & KernelShap & 0.439 & 0.169 \\ 
                    \hline
                    ConvTran & random & 0.444 & 0.246 \\

                    \hline
                    MiniRocket & \textbf{F. Ablation} & \textbf{0.449} & 0.187 \\ 
                    MiniRocket & F. Permutation & 0.440 & \textbf{0.205} \\ 
                    MiniRocket & SHAP & 0.416 & 0.177 \\ 
                    MiniRocket & KernelShap & 0.401 & 0.152 \\ 
                    \hline
                    MiniRocket & random & 0.408 & 0.241 \\
                    \hline
                \end{tabular}
        \end{minipage} &
    
        \begin{minipage}{.55\linewidth}
                \begin{tabular}{|c|c|c|c|}
                    \hline
                    Model  &  XAI method   & $AUC\tilde{S}_{top}$   & $F1\mathcal{S}$ \\
                    \hline
                    ResNet & \textbf{DeepLiftShap} & \textbf{0.380} & \textbf{0.162} \\
                    ResNet & F. Ablation & 0.376 & 0.162  \\
                    ResNet & DeepLift & 0.375 & 0.153 \\
                    ResNet & Int. Gradients & 0.375 & 0.152 \\
                    ResNet & F. Permutation & 0.374 & 0.182 \\
                    ResNet & SHAP & 0.373 & 0.159 \\
                    ResNet & GradientShap & 0.373 & 0.152 \\
                    ResNet & Saliency & 0.368 & 0.156 \\
                    ResNet & KernelShap & 0.367 & 0.152 \\
                    \hline
                    ResNet & random & 0.371 & 0.233 \\
                
                    \hline
                    dResNet & \textbf{SHAP} & \textbf{0.416} & 0.164 \\ 
                    dResNet & F. Ablation & 0.414 & 0.175 \\ 
                    dResNet & F. Permutation & 0.413 & 0.197 \\ 
                    dResNet & dCAM & 0.410 & \textbf{0.243} \\ 
                    dResNet & KernelShap & 0.401 & 0.164 \\
                    \hline
                    dResNet & random & 0.416 & 0.242 \\
                    \hline
                \end{tabular}
        \end{minipage} 
    \end{tabular}
\caption{ $AUC\tilde{S}_{top}$  and $F1\mathcal{S}$ for different attribution methods sorted by $AUC\tilde{S}_{top}$ , grouped by different classifiers. The whole pipeline (train the classifier, run  XAI methods, run InterpretTime with no data augmentation) was performed using our synthetic dataset implementation. All metrics, for each classifers, are very close to the random accuracy.}
\label{tab:InterpretTime_baseline}
\end{table}

\newpage

\subsection*{InterpretTime results for ECG dataset}
\begin{table}[ht!]
                \begin{tabular}{|c|c|c|c|}

                    \hline
                    Model  &  XAI method  & \multicolumn{2}{|c|}{  $AUC\tilde{S}_{top}$ }   \\
                    \hline 
                    \multicolumn{2}{|c|}{} & Point-wise & Chunk-wise  \\

                     \hline
                    ConvT. & Int. Gradients & 0.566  & /  \\ 
                    ConvT. & GradientShap & 0.519  &  / \\ 
                    ConvT. & SHAP & 0.472 & 0.585 \\ 
                    ConvT. & F. Ablation & 0.436 & 0.522  \\ 
                    ConvT. & F. Permutation & 0.376 & 0.370 \\ 
                    ConvT. & DeepLiftShap & 0.340   & / \\ 
                    ConvT. & DeepLift & 0.340  & / \\ 
                    ConvT. & KernelShap & 0.114 & 0.292 \\ 
                    ConvT. & Saliency & 0.058  & / \\ 
                    \hline 
                    ConvT. & Random & 0.077 & 0.125  \\ 
                     
                    \hline
                    MiniR. & SHAP & 0.800 & 0.679  \\ 
                    MiniR. & F. Ablation & 0.701 & 0.751  \\ 
                    MiniR. & F. Permutation & 0.550  & 0.620  \\ 
                    MiniR. & KernelShap & 0.390 & 0.486 \\ 
                    \hline
                    MiniR. & Random & 0.357 & 0.336 \\ 
                    \hline
                    
                    ResN. & SHAP & 0.482 & 0.470  \\ 
                    ResN. & DeepLift & 0.427  & / \\ 
                    ResN. & Int. Gradients & 0.416  & / \\ 
                    ResN. & F. Ablation & 0.356 &  0.379 \\ 
                    ResN. & DeepLiftShap & 0.334 &  / \\ 
                    ResN. & F. Permutation & 0.307 &  0.321 \\ 
                    ResN0. & GradientShap & 0.306 &  / \\ 
                    ResN. & KernelShap & 0.197 & 0.259  \\ 
                    ResN. & Saliency & 0.149 &  / \\ 
                    \hline 
                    ResN. & Random & 0.173 & 0.164   \\ 
                    \hline 

                    dResN. & SHAP & 0.351  & 0.412  \\ 
                    dResN. & F. Ablation & 0.336 &  0.373 \\ 
                    dResN. & F. Permutation & 0.316 & 0.333  \\ 
                    dResN. & dCAM & 0.314 &  /\\ 
                    dResN. & KernelShap & 0.263 & 0.287 \\ 
                    \hline
                    dResN. & Random & 0.253  & 0.221  \\                 
                    \hline  
                \end{tabular}
 
\caption{ InterpretTime results using ECG dataset sorted by $AUC\Tilde{{S}}$ for point-wise case. All distributions were included apart from Global Gaussian and normal mean using MiniRocket chunk-wise. Focusing on the 2 models we applied both gradient and perturbation-based methods, for ConvTran the best score is SHAP chunk-wise followed by integrated Gradient, while for ResNet the two best scores are both from SHAP (point and chunk-wise). }
\label{tab:ECG_results}
\end{table}

\newpage

\subsection*{Different masks for MP dataset}
\begin{table}[h]

            \centering
            \begin{tabular}{|c|c|c|c|c|c|c|}
            \hline
            XAI Method & \multicolumn{2}{|c|}{normal distribution}  &  \multicolumn{2}{|c|}{local Gaussian}   &  \multicolumn{2}{|c|}{global Gaussian}  \\
            \hline
             & $AUC\tilde{S}_{top}$ & $F1\mathcal{S}$ & $AUC\tilde{S}_{top}$ & $F1\mathcal{S}$ & $AUC\tilde{S}_{top}$ & $F1\mathcal{S}$ \\ 
             
            \hline
            FeatureAblation & 0.776 & \textbf{0.299} & 0.608 & 0.244 & \textbf{0.699} & \textbf{0.279} \\ 
            FeaturePermutation & 0.653 & 0.249 & \textbf{0.654} & \textbf{0.262} & 0.633 & 0.246 \\ 
            KernelShap & 0.681 & 0.220 & 0.484 & 0.147 & 0.656 & 0.218 \\ 
            SHAP & \textbf{0.871} & 0.289 & 0.549 & 0.213 & 0.674 & 0.243 \\ 
            \hline     
            Random & 0.522 & 0.165 & 0.289 & 0.106 & 0.565 & 0.182 \\    
            \hline

            \end{tabular}
            \caption{InterpretTime scores for 3 different mask using MP dataset and ConvTran classifier, 15 chunks. Each of the masks provides a different best model.}
            
\end{table}

Accuracy vs percentage of corrupted time series ($\tilde{N}$) for three different masks and relatively different behaviors for ConvTran classifier, 15 chunks, MP dataset

\begin{figure}[ht!]
    \centering
    \includegraphics[width=0.8\textwidth] {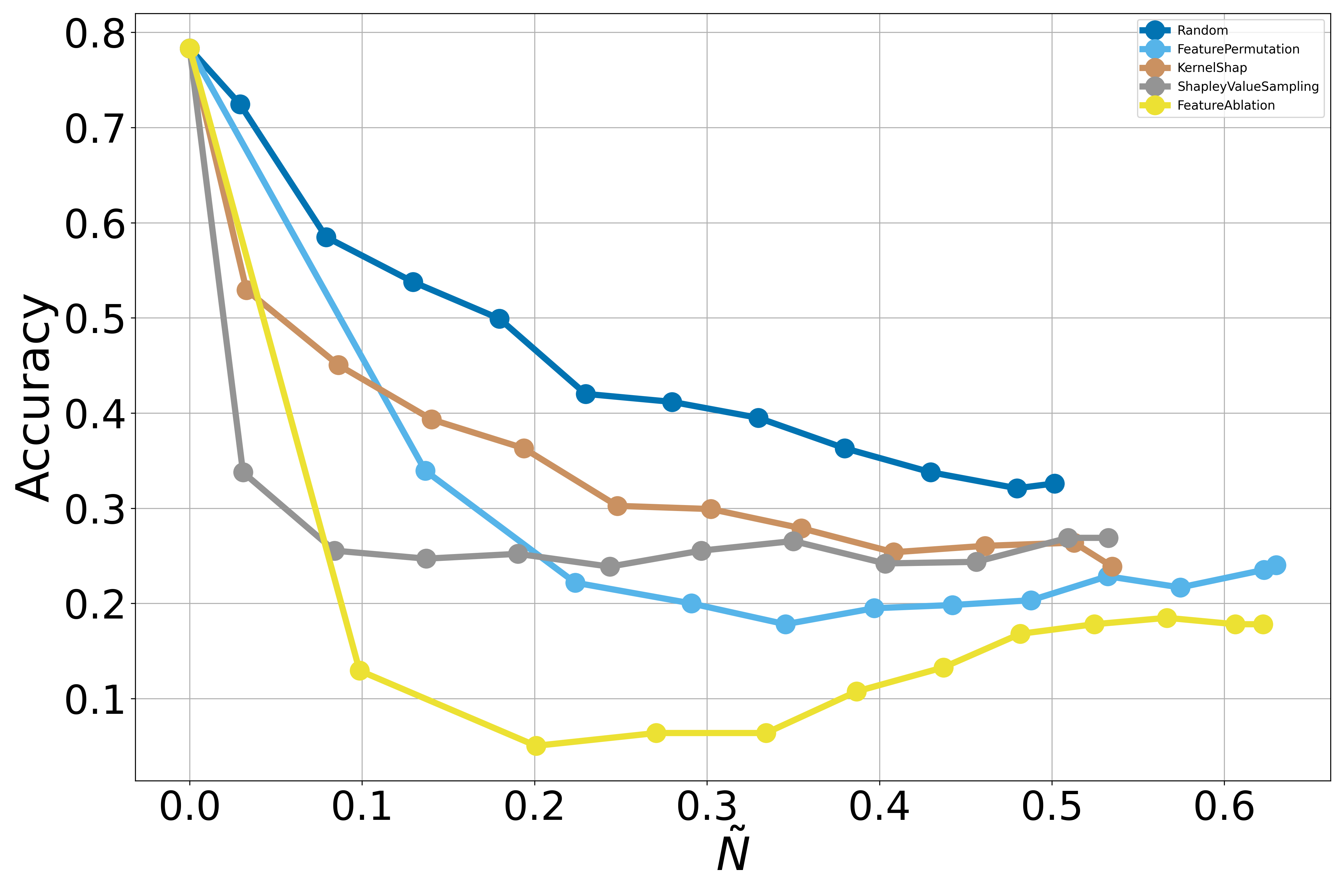}
    \caption{Global Gaussian}
\end{figure}

\begin{figure}[ht!]
    \centering
    \includegraphics[width=0.8\textwidth] {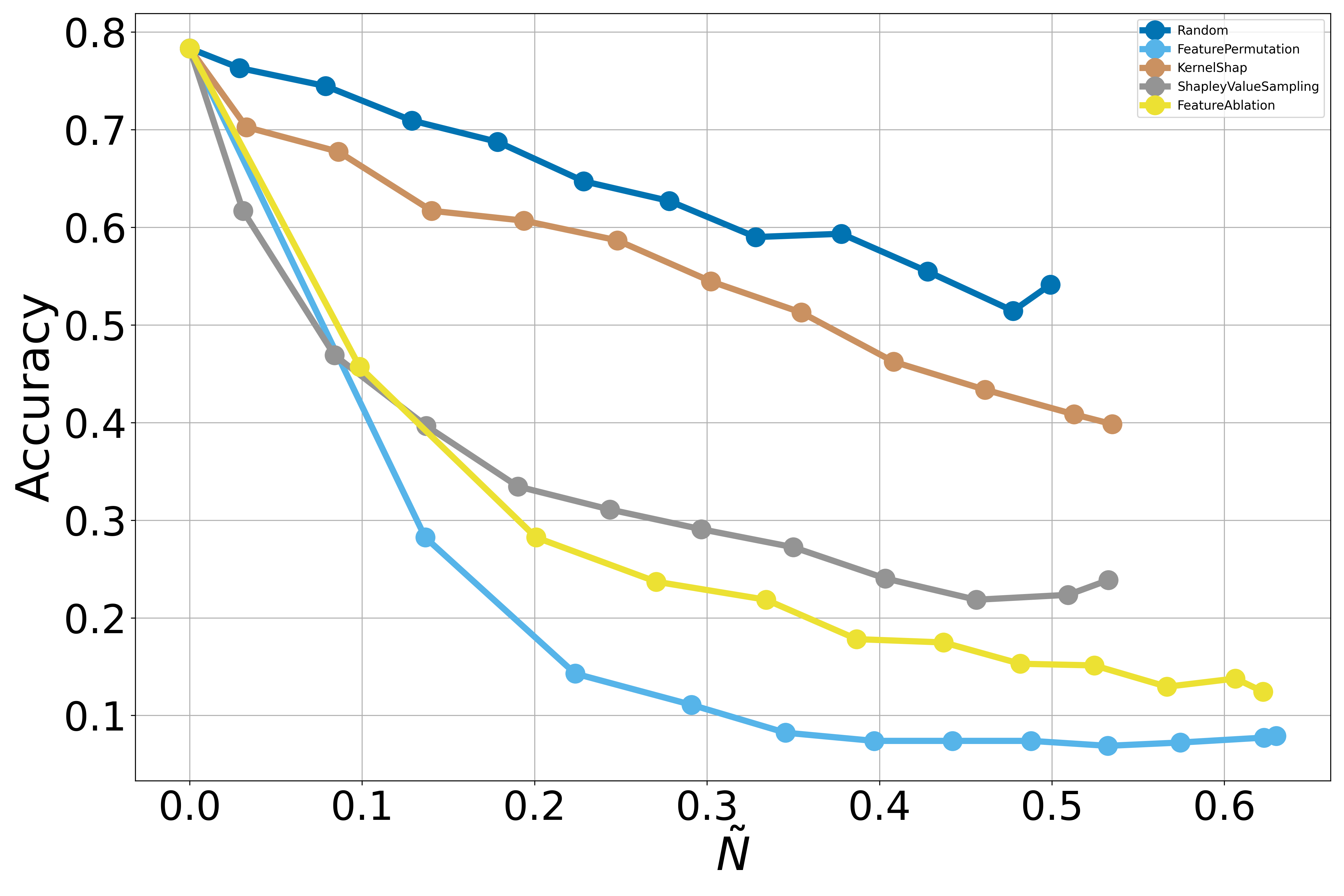}
    \caption{Local Gaussian}
\end{figure}

\begin{figure}[ht!]
    \centering
    \includegraphics[width=0.8\textwidth] {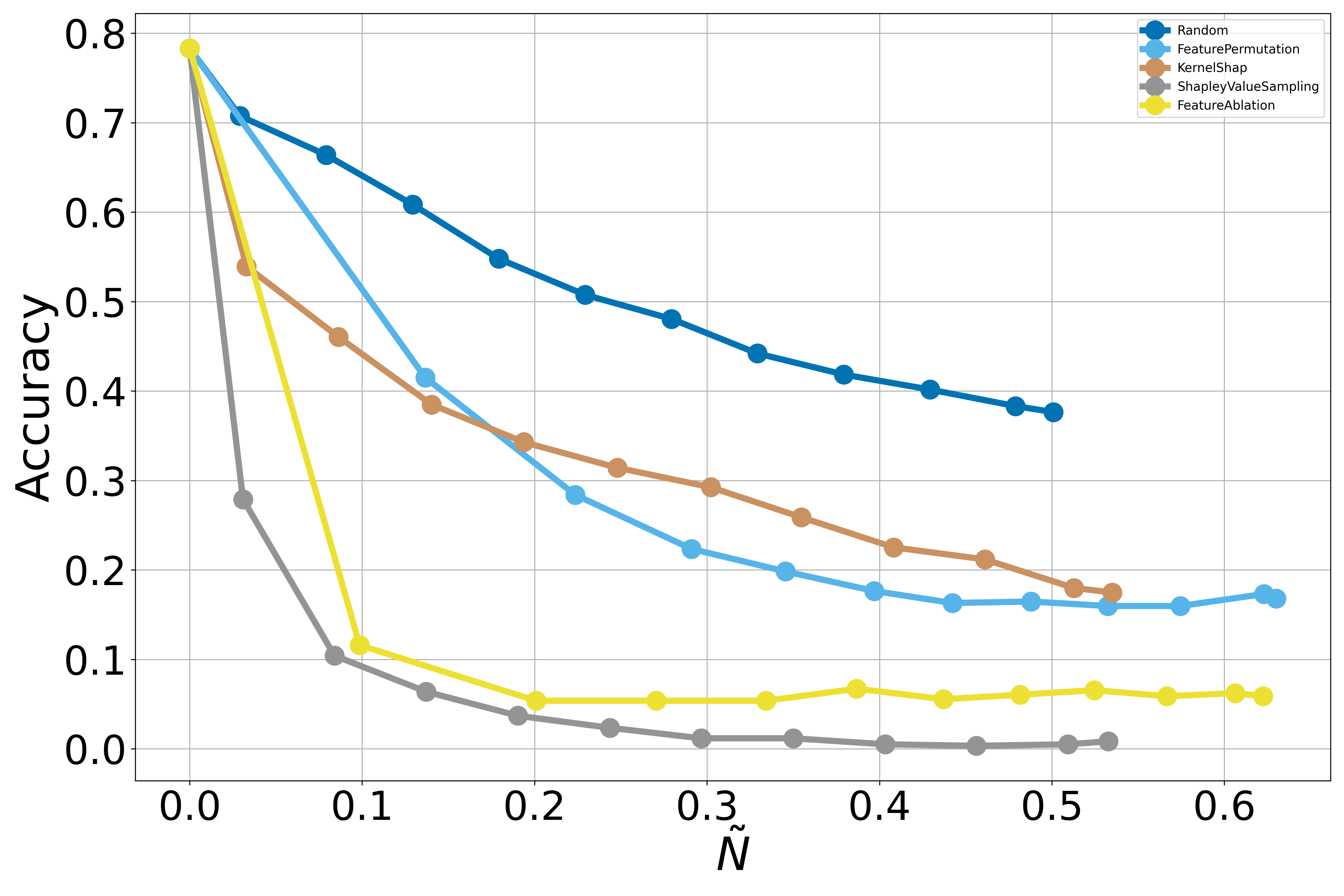}
    \caption{Normal distribution}
\end{figure}

\clearpage
\newpage

\subsection*{Accuracy vs percentage of corrupted time series for synthetic data}

Accuracy drop vs percentage of values substituted ($\tilde{N}$) curves for synthetic data using the original InterpretTime methodology (i.e. using values sampled from normal distribution as a mask, but no train data augmentation). All explanation methods applied to all classifiers converge very fast to random accuracy ($0.5$ since this is a binary problem) because they classify all instances as negative ones.

\begin{figure}[ht!]
    \centering
    \includegraphics[width=0.8\textwidth] {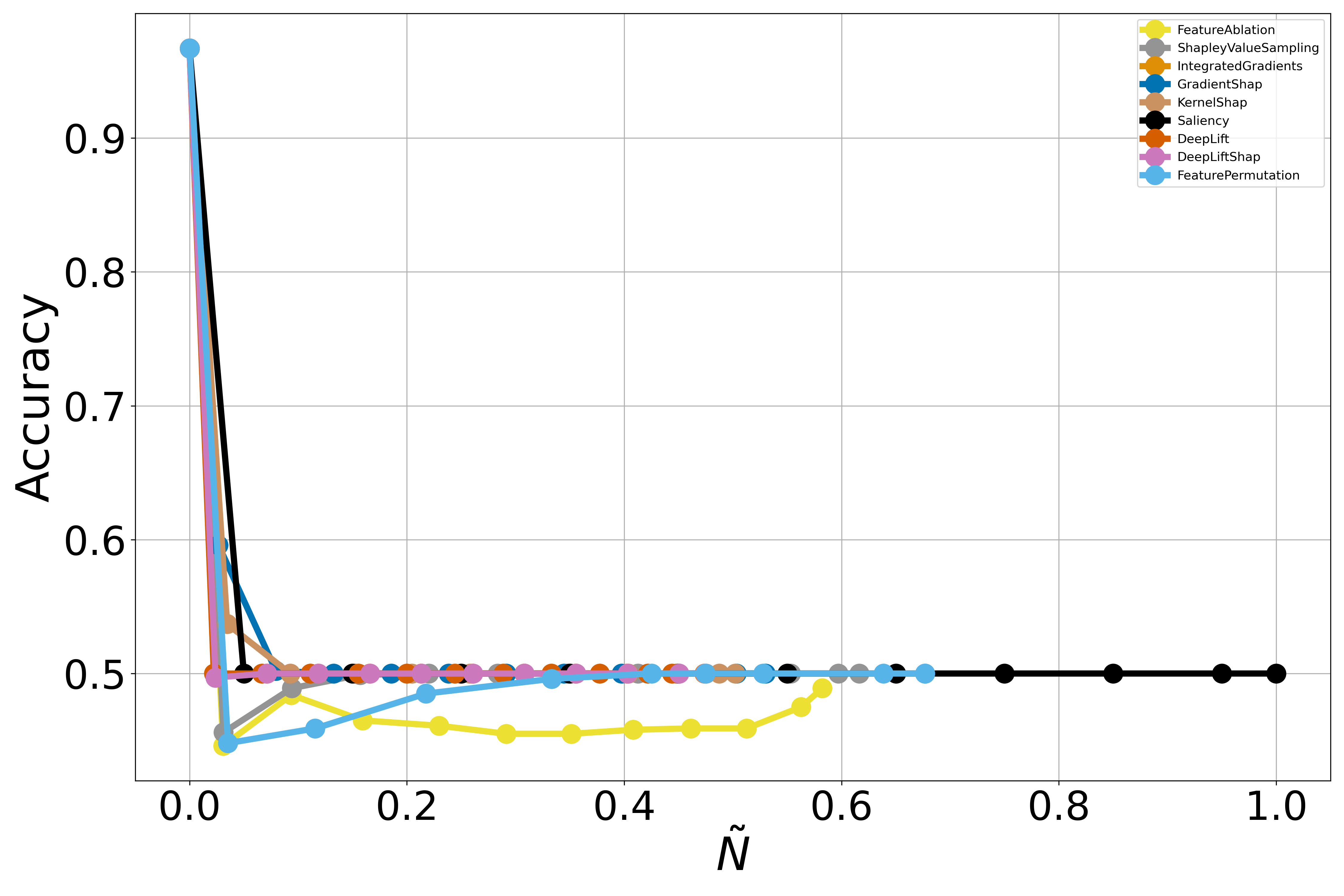}
    \caption{ConvTran}
\end{figure}

\begin{figure}[ht!]
    \centering
    \includegraphics[width=0.8\textwidth] {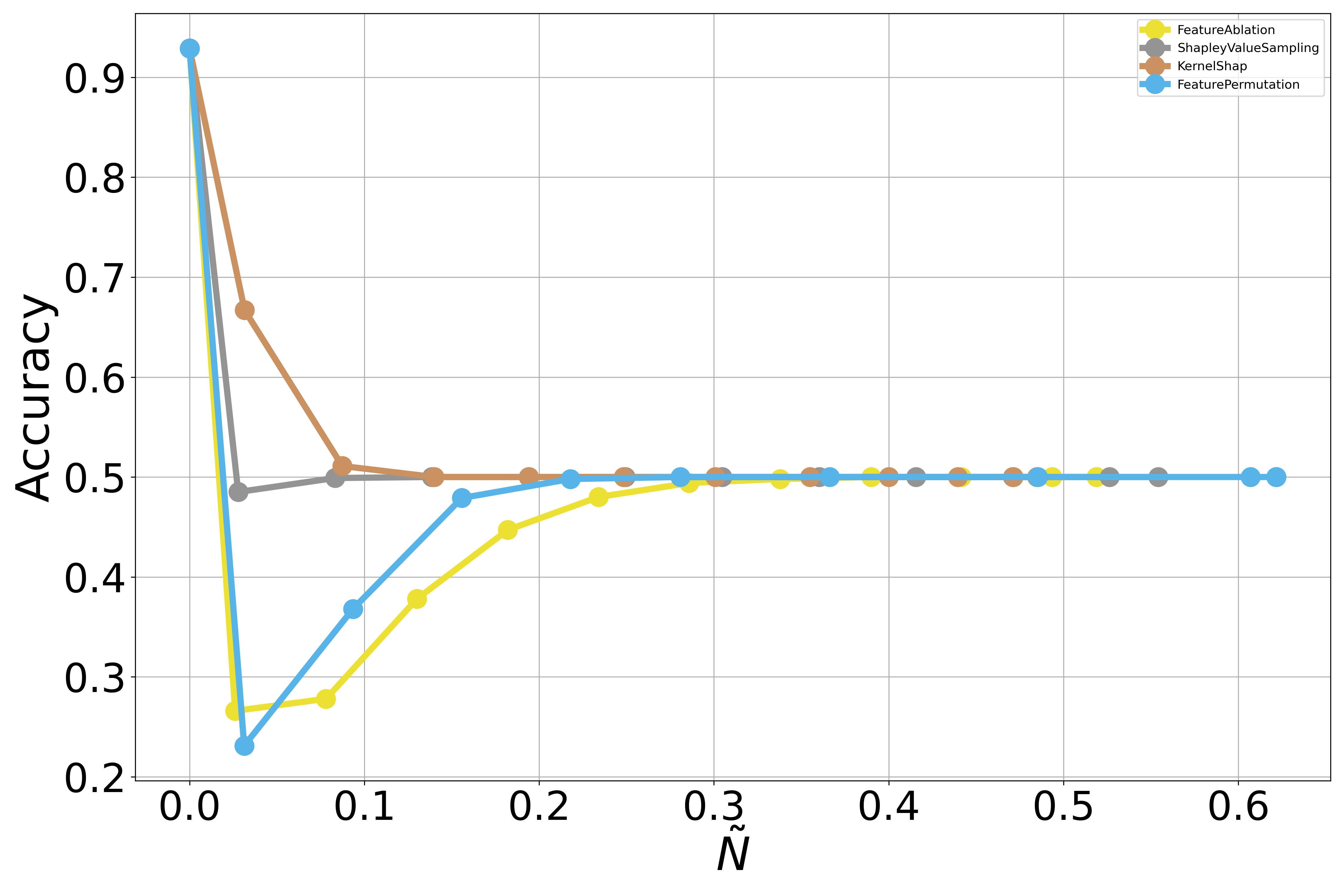}
    \caption{MiniRocket}
\end{figure}

\begin{figure}[ht!]
    \centering
    \includegraphics[width=0.8\textwidth] {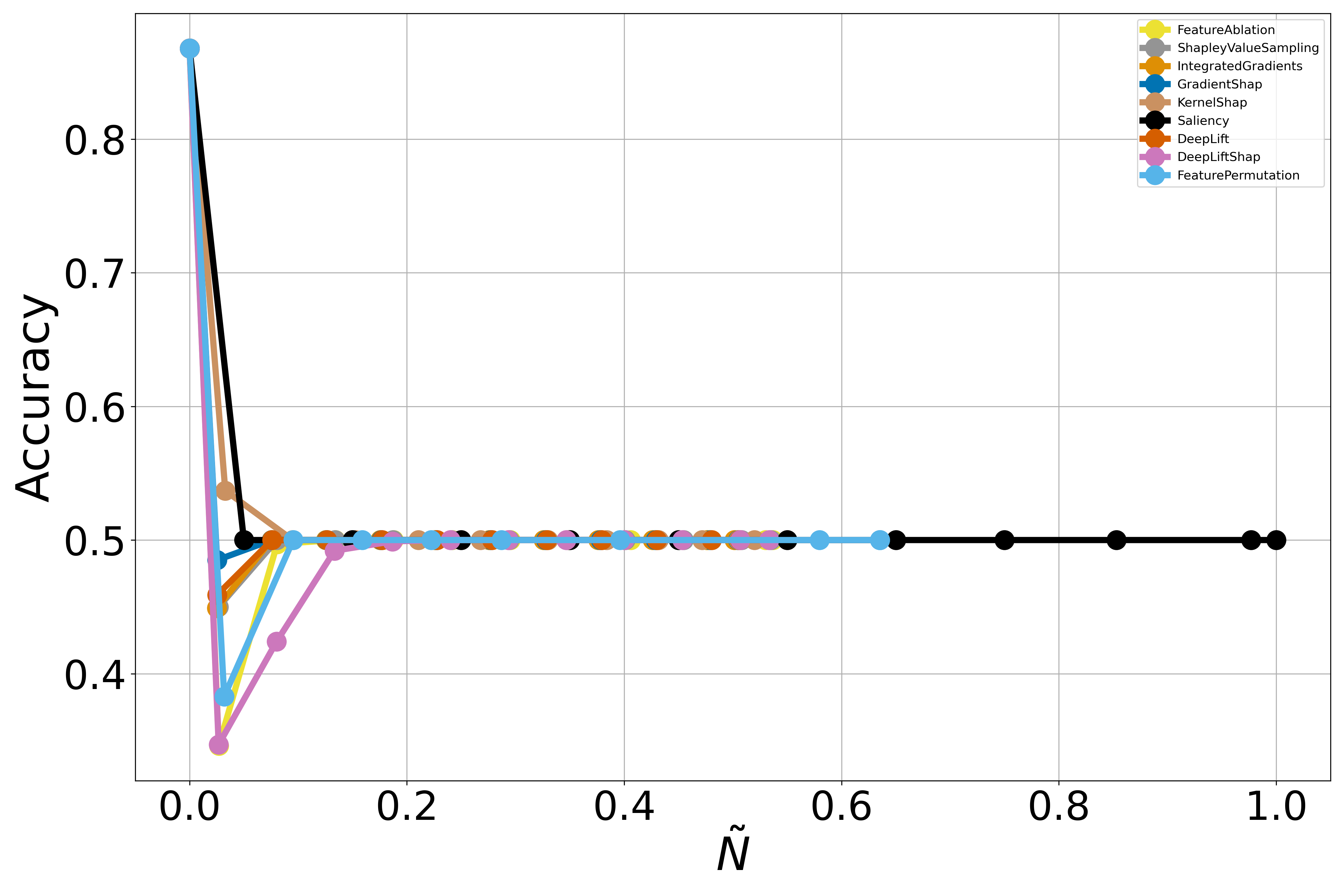}
    \caption{ResNet}
\end{figure}

\begin{figure}[ht!]
    \centering
    \includegraphics[width=0.8\textwidth] {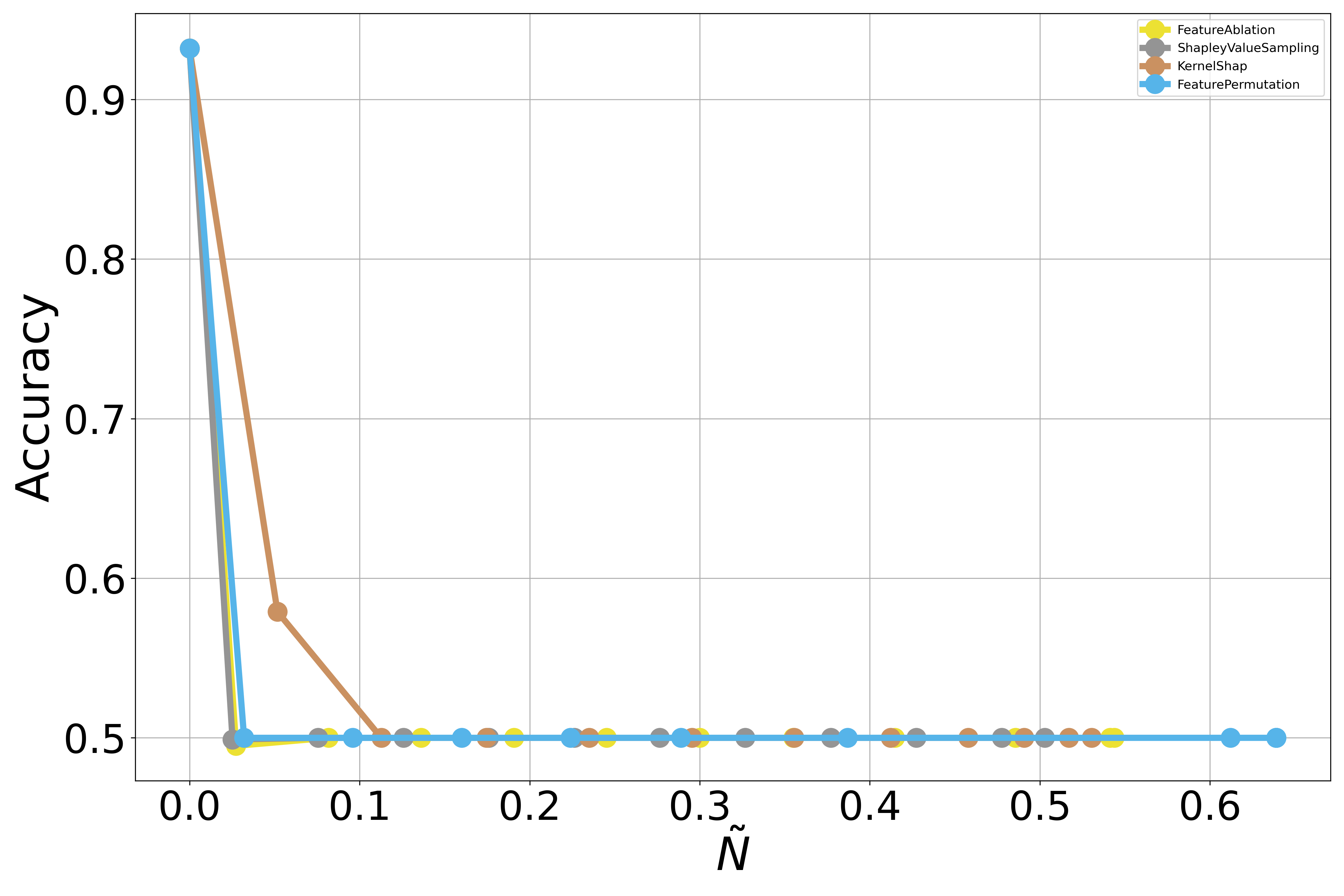}
    \caption{dResNet}
\end{figure}


\clearpage

\subsection*{Chunking results}
\begin{table}[ht]
    \centering
    \begin{tabular}{|c|c|c|c|c|c|c|}
    \hline
     Dataset & Classifier & XAI method & \multicolumn{4}{c|}{  $AUC\Tilde{{S}}$ for Number of chunks}  \\
     \hline
        & &  & 5 & 10 & 15 & 20 \\
    \hline
    
    MP & ConvTran & SHAP & 0.743 & \textbf{0.757} & 0.745 & 0.750 \\ 
    MP & ConvTran & FeatureAblation & 0.686 & \textbf{0.710} & 0.701 & 0.708 \\ 
    MP & ConvTran & FeaturePermutation & 0.633 & 0.641 & \textbf{0.648} & 0.628 \\ 
    MP & ConvTran & KernelShap & \textbf{0.608} & 0.603 & 0.597 & 0.586 \\ 
    \hline
    MP & ConvTran & Random & 0.485 & 0.477 & 0.461 & 0.464 \\ 
    
    \hline

    CMJ &    MiniRocket & KernelShap & \textbf{0.582} & 0.457 & 0.454 & 0.409 \\ 
    CMJ &    MiniRocket & FeatureAblation & 0.746 & \textbf{0.760} & 0.749 & 0.733 \\ 
    CMJ &    MiniRocket & FeaturePermutation & \textbf{0.472} & 0.440 & 0.446 & 0.430 \\ 
   CMJ &     MiniRocket & SHAP & 0.758 & \textbf{0.762} & 0.759 & 0.755 \\ 
        \hline 
    CMJ &    MiniRocket & Random & \textbf{0.294} & 0.221 & 0.223 & 0.213 \\ 
    \hline
    synthetic & ResNet & FeatureAblation & 0.542 & 0.601 & \textbf{0.620} & 0.571 \\ 
    synthetic & ResNet & SHAP & 0.637 & 0.677 & \textbf{0.691} & 0.528 \\ 
    synthetic & ResNet & KernelShap & 0.358 & \textbf{0.370} & 0.352 & 0.327 \\ 
    synthetic & ResNet & FeaturePermutation & 0.444 & 0.485 & \textbf{0.504} & 0.489 \\ 
    \hline
    synthetic & ResNet & Random & 0.161 & 0.162 & 0.153 & 0.203 \\ 

    \hline
    \end{tabular}
    \caption{ Detailed performances for 3 classifiers, each one using a different dataset. For each of these 3 groups, the best performance was achieved either using 10 or 15 chunks.}
    \label{tab:my_label}
\end{table}

\end{document}